\documentclass{fairmeta}
\usepackage{horizonlab}

\usepackage{amsmath,amsfonts,bm}

\def\eqref#1{equation~\ref{#1}}

\def\1{\bm{1}}

\DeclareMathAlphabet{\mathsfit}{\encodingdefault}{\sfdefault}{m}{sl}
\SetMathAlphabet{\mathsfit}{bold}{\encodingdefault}{\sfdefault}{bx}{n}

\usepackage{cleveref}

\newcommand{\paperlogos}{%
    \includegraphics[height=0.92cm]{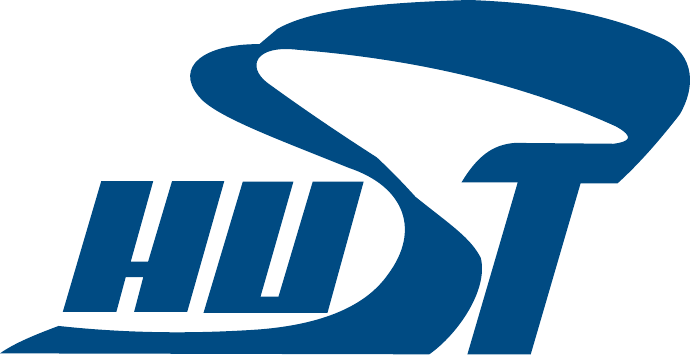}%
    \hspace{0.3cm}%
    \includegraphics[trim=0 30 26.64 34.08,clip,height=0.92cm]{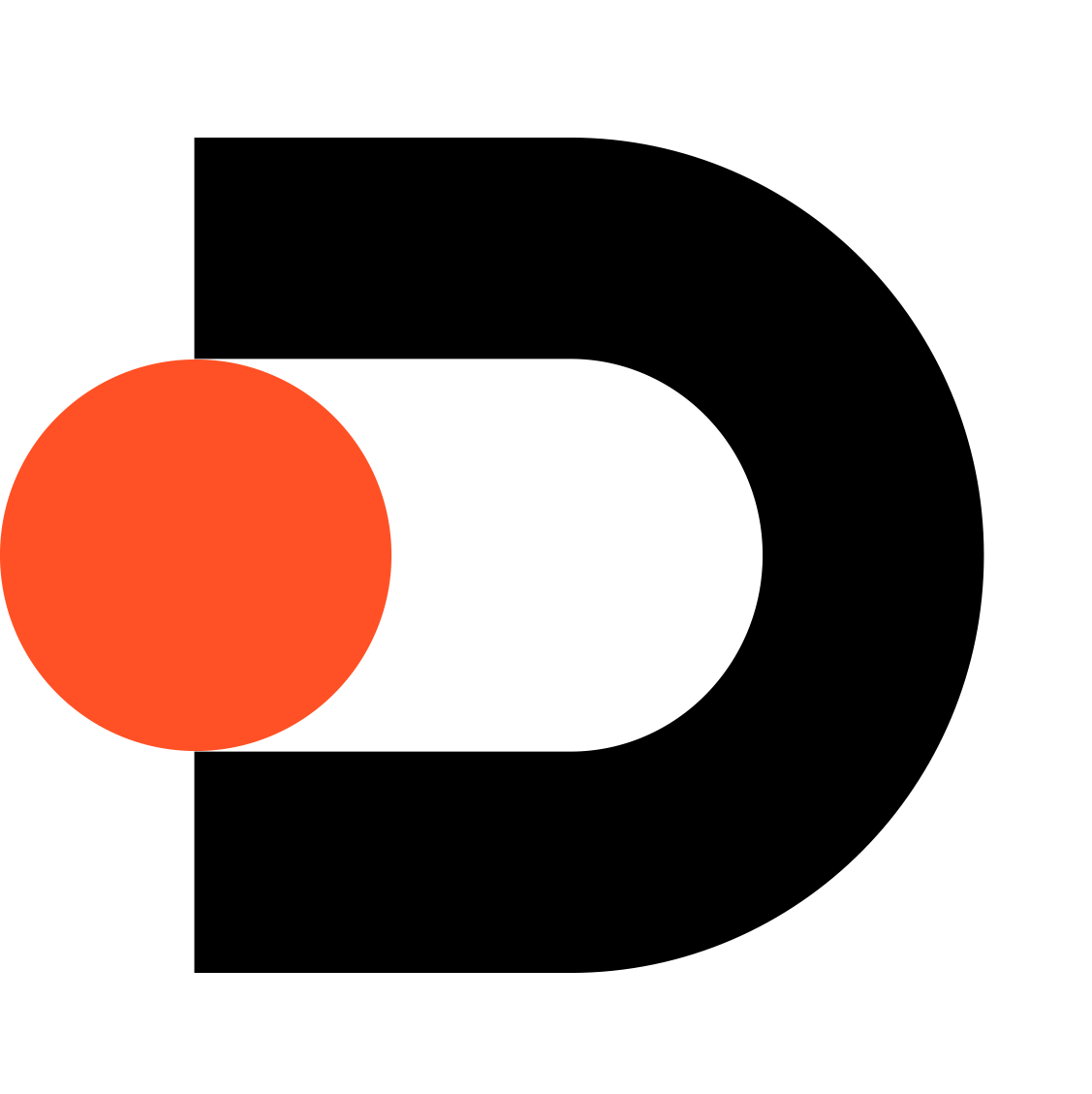}%
    \hspace{0.36cm}%
    \includegraphics[trim=76 21 82 19,clip,height=0.92cm]{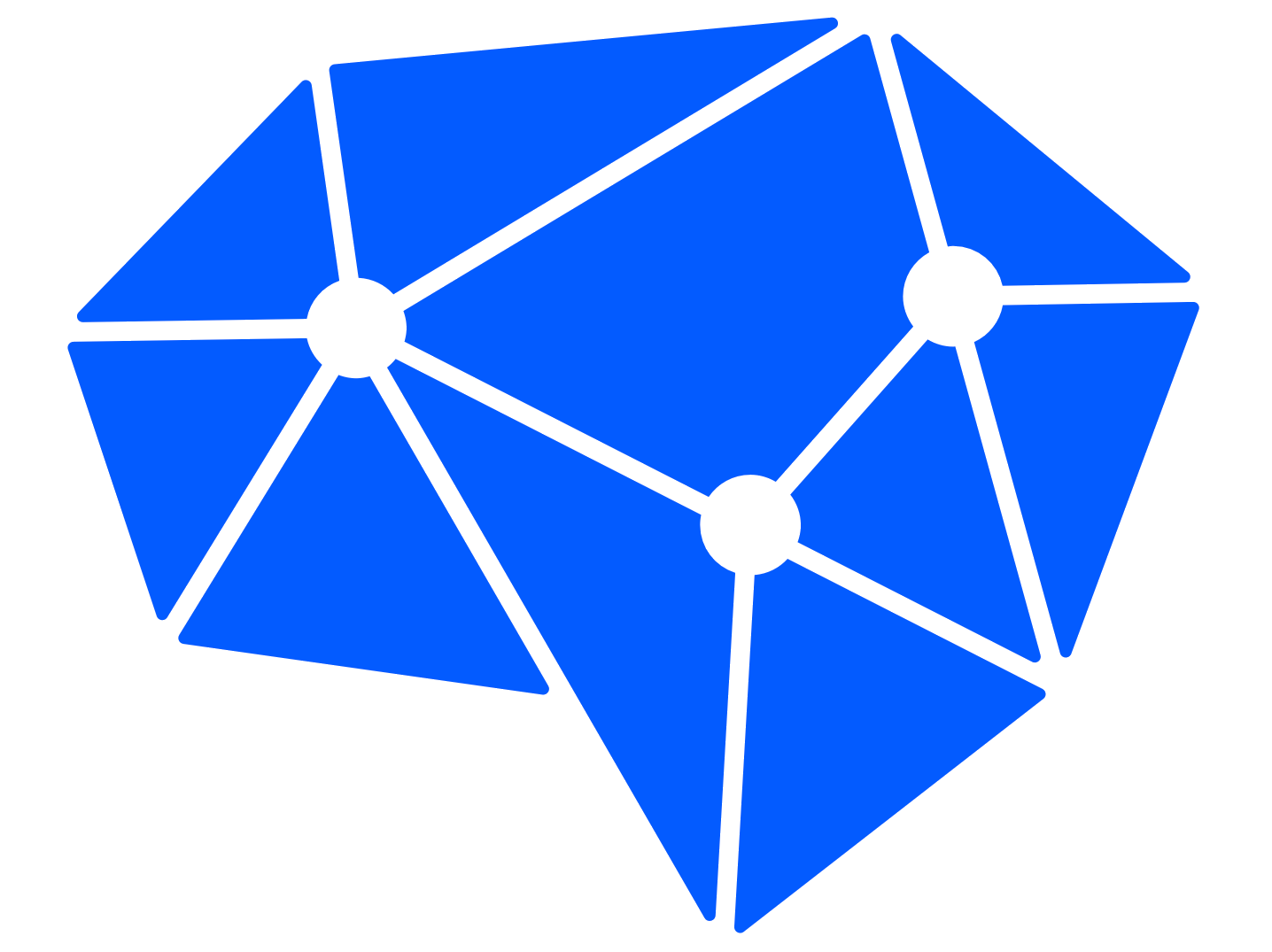}%
}
\definecolor{paperblue}{rgb}{0.13,0.372,0.76}

\crefname{figure}{fig.}{figs.}
\crefname{table}{tab.}{tabs.}
\crefname{equation}{eq.}{eqs.}
\crefname{section}{sec.}{secs.}
\crefname{subsection}{sec.}{secs.}
\crefname{subsubsection}{sec.}{secs.}
\crefname{appendix}{app.}{apps.}

\Crefname{figure}{Fig.}{Figs.}
\Crefname{table}{Tab.}{Tabs.}
\Crefname{equation}{Eq.}{Eqs.}
\Crefname{section}{Sec.}{Secs.}
\Crefname{subsection}{Sec.}{Secs.}
\Crefname{subsubsection}{Sec.}{Secs.}
\Crefname{appendix}{App.}{Apps.}

\title{\calmfont{Rethinking Representations for World-Action Modeling}}

\author[1,*]{\calmfont{Haoyi Jiang}}
\author[3,\dagger]{\calmfont{Liu Liu}}
\author[3]{\calmfont{Xinjiang Wang}}
\author[4]{\calmfont{Zhihao Sun}}
\author[3]{\calmfont{Zequn Chen}}
\author[5]{\calmfont{Sen Wang}}
\author[3]{\calmfont{Xinjie Wang}}
\author[3]{\calmfont{Xia Chen}}
\author[1]{\calmfont{Jingfeng Yao}}
\author[1]{\calmfont{Weiheng Zhao}}
\author[1]{\calmfont{Shanglin Yuan}}
\author[3]{\calmfont{Zhizhong Su}}
\author[2]{\calmfont{Wei Sui}}
\author[1]{\calmfont{Wenyu Liu}}
\author[1,\ddagger]{\calmfont{Xinggang Wang}}

\affiliation[1]{\calmfont{Huazhong University of Science \& Technology}}
\affiliation[2]{\calmfont{D-Robotics}}
\affiliation[3]{\calmfont{Horizon Robotics}}
\affiliation[4]{\calmfont{Fudan University}}
\affiliation[5]{\calmfont{Xi'an Jiaotong University}}
\contribution[*]{Intern at D-Robotics}
\contribution[\dagger]{Project leader}
\contribution[\ddagger]{Corresponding author}
\code{\url{https://github.com/hustvl/ReWAM}}
\metadata[Contact]{Haoyi Jiang at \email{haoyi\_jiang@hust.edu.cn}}
\correspondence{Xinggang Wang at \email{xgwang@hust.edu.cn}}

\begin{document}
\abstract{
    World-action models jointly learn robot policies and predict future observations, making the representation space an interface between control and prediction. We study the design of this space through controlled comparisons, finding that neither reconstruction fidelity nor pre-trained perceptual features alone ensure effective policy learning. These findings motivate ReWAM, a representation-centric world-action model built on pre-trained DINO features. Feature Calibration and a Temporal Representation Bottleneck organize these features into compact world states suited to dynamics modeling. Action-Grounded Representation Shaping routes only action-loss gradients to the bottleneck, thereby letting the policy shape what the representation encodes while the world model learns how it evolves. Without generative video pre-training, ReWAM achieves 93.6\% success on RoboTwin 2.0. On RoboDojo, it achieves an average score of 12.29 and a success rate of 8.28\% using approximately 600 hours of embodied pre-training data.
}

\maketitle

\section{Introduction}

World-action models (WAMs) jointly learn robot policies and predict how the world evolves. Many recent WAMs~\citep{Motus,LingBot-VA,DreamZero,Fast-WAM} build on pre-trained video generators and predict future observations in the latent space of a video variational autoencoder (Video-VAE). These models inherit dynamics priors from large-scale generative video pre-training, but their latent spaces are typically optimized for visual reconstruction. Such spaces may preserve fine-grained appearance details without explicitly prioritizing information relevant to control.

The role of world prediction extends beyond producing explicit visual rollouts. Fast-WAM~\citep{Fast-WAM} retains video co-training while removing future-video denoising at deployment, suggesting that prediction can benefit policies through joint training. This raises a fundamental question: \textbf{How should the representation space for world-action modeling be defined?}

Representation-space WAMs use pre-trained perceptual features or learned visual-action tokenizers~\citep{LDA-1B,DexWorldModel,RepWAM}. Comparative studies show that reconstruction fidelity alone is insufficient to assess control utility~\citep{semantic-wm,ActRelevantLats}. Our experiments support this distinction: without generative video pre-training, raw DINO features outperform Video-VAE latents on RoboTwin 2.0~\citep{RoboTwin}, despite lower pixel-reconstruction SSIM. Reconstruction-oriented latents therefore do not by themselves establish effective representations for policy learning.

Perceptual features are also not necessarily well suited to dynamics modeling: frame-wise features do not explicitly encode temporal variation, and their statistics may be poorly matched to diffusion training. Calibrating DINO features through multi-layer aggregation, normalization, and a representation-aware diffusion noise schedule improves average success on RoboTwin 2.0 by $12.20$ percentage points. Representation design thus extends beyond the choice of encoder. It must also address how features are prepared for prediction, organized over time, and adapted to the needs of control.

\begin{figure}[t]
    \centering
    \includegraphics[width=\linewidth]{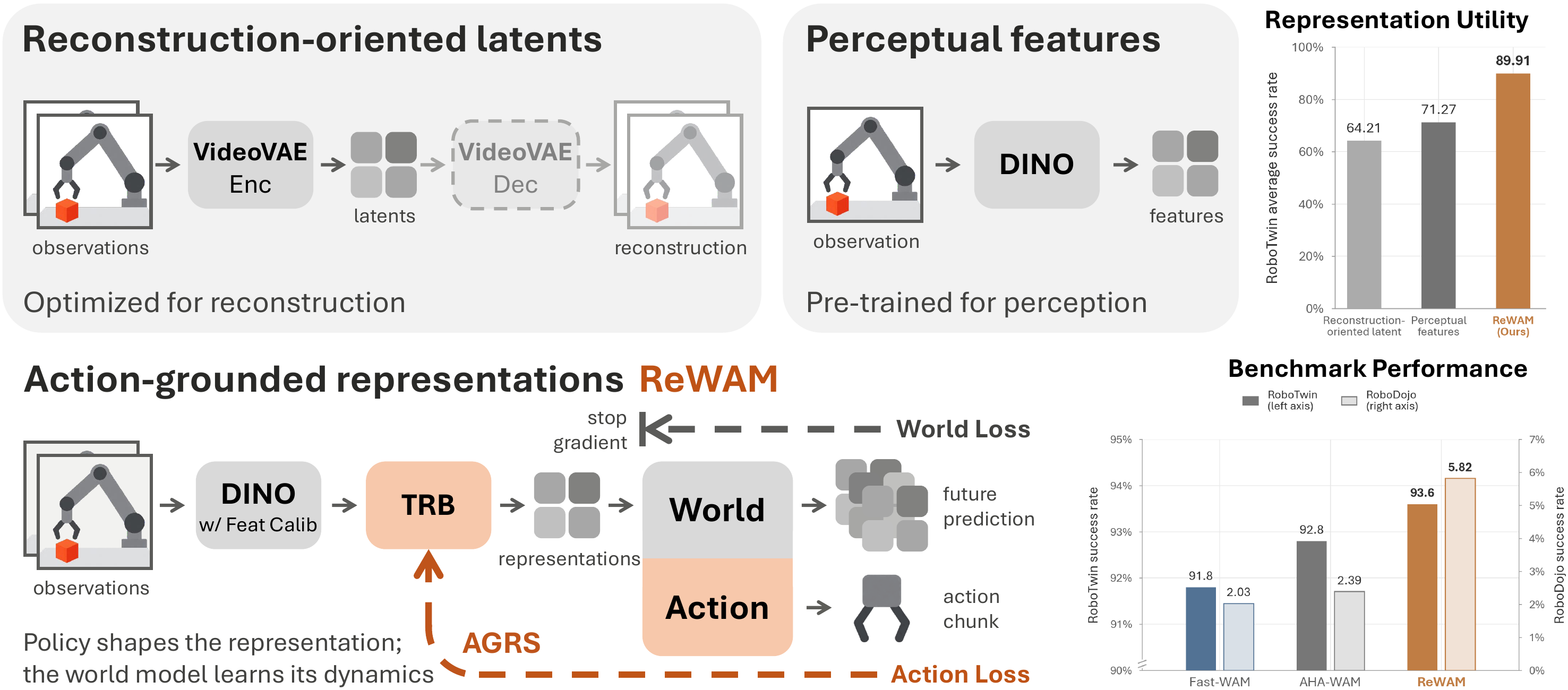}
    \caption{\textbf{Overview of representations for world-action modeling.} Top-left: Video-VAE latents are optimized for visual reconstruction. Top-right: Frozen DINO features inherit priors from perceptual pre-training. Bottom: ReWAM uses a Temporal Representation Bottleneck (TRB) to form compact world states from calibrated DINO features. Action-Grounded Representation Shaping (AGRS) trains the TRB through asymmetric routing of action-loss gradients, while the world model learns state transitions.}
    \label{fig:overview}
\end{figure}

We propose a division of roles: \textbf{the policy shapes what the representation encodes, while the world model learns how that representation evolves.} We introduce \textbf{ReWAM}, a representation-centric WAM that implements this principle through Action-Grounded Representation Shaping (AGRS). Feature Calibration prepares frozen DINO features for diffusion modeling, and a Temporal Representation Bottleneck (TRB) organizes them into compact world states. During joint training, AGRS routes only action-loss gradients to the bottleneck, adapting the representation to action prediction. The world model learns how this representation evolves without updating it through its own prediction loss. \Cref{fig:overview} provides an overview of ReWAM.

Without generative video pre-training, ReWAM achieves 93.6\% success in both the clean and random settings of RoboTwin 2.0 and leads the compared methods without embodied pre-training on RoboDojo~\citep{RoboDojo}. Embodied pre-training further improves its RoboDojo average score to 12.29 and success rate to 8.28\%. Together with real-robot evaluations, these findings highlight a promising path for world-action modeling based on pre-trained perceptual priors and representations shaped by the demands of control, without relying on generative video pre-training.

Our contributions are threefold:

\begin{itemize}
    \item We show through controlled comparisons that reconstruction fidelity alone is insufficient for choosing WAM representations, and that perceptual features benefit from calibration for dynamics modeling.
    \item We introduce ReWAM, which lets the action objective shape a compact representation space for joint prediction and control.
    \item We demonstrate the effectiveness of this design without generative video pre-training across two simulation benchmarks and real-robot tasks.
\end{itemize}

\section{Related Work}

\subsection{World-Action Models}

Unified video and action models couple dynamics prediction with policy learning~\citep{UWM,UVA}. Recent systems build on video generation through joint denoising~\citep{DreamZero} or interactions between world and action streams~\citep{Motus,LingBot-VA,MotuBrain}. RxBrain~\citep{RxBrain} and Cosmos 3~\citep{Cosmos3} extend this framework to broader multimodal understanding and generation.

Other approaches expose intermediate generative features to the policy~\citep{mimic-video,DiT4DiT,ImageWAM}. Fast-WAM~\citep{Fast-WAM} removes future-video denoising at deployment, while Faster-WAM~\citep{Faster-WAM} studies sparse interactions between video and action prediction under distribution shift. These studies motivate examining how the shared representation interface supports policy learning. Complementary work enriches visual prediction with geometric, motion, and semantic supervision~\citep{X-WAM,WAM4D,DreamWAM,EgoWAM}, or aligns policy features with future visual representations~\citep{FLARE,FRAPPE}.

Representation-space approaches use pre-trained visual features for prediction and control. LDA-1B~\citep{LDA-1B} and DexWorldModel~\citep{DexWorldModel} directly predict DINO features, while VLA-JEPA~\citep{VLA-JEPA} and Being-H0.7~\citep{Being-H0.7} use future observations to supervise latent representations that guide action generation. LaWAM~\citep{LaWAM} predicts latent visual subgoals, and GAM~\citep{GAM} uses geometric foundation features for prediction and action decoding. Learned visual-action tokenizers adapt the interface between perception and control~\citep{RepWAM,LingBot-VA2.0,LiLa-WAM}, while residual latent actions model changes in DINO features~\citep{RLA}. ReWAM builds on this line of work by calibrating and temporally organizing pre-trained features, then using asymmetric gradient routing to adapt the prediction space to the policy objective during joint world-action training.

\subsection{Representations for Prediction and Generation}

Predictive learning in visual feature spaces supports representation learning and world modeling, as explored by V-JEPA models~\citep{V-JEPA2,V-JEPA2.1}, DINO-world~\citep{DINO-World}, and Spa3R~\citep{Spa3R}.

Representation design also matters for generation. REPA~\citep{REPA,REPA-E} and VA-VAE~\citep{VA-VAE} align denoiser features and autoencoder latents, respectively, with pre-trained representations. The representation autoencoder (RAE)~\citep{RAE,RAEv2} uses pre-trained encoders to define generative latent spaces, with subsequent work extending this approach to video generation~\citep{VideoRAE,V-RAE} and interactive world modeling~\citep{MIRA}.

Probing studies and comparative analyses examine what representations encode and how they support spatial understanding, dynamics prediction, and control~\citep{VidFM-3D-Probe,ActRelevantLats,semantic-wm,NanoWM}. ReWAM addresses the complementary problem of how to construct and adapt a representation space for joint world prediction and policy learning.

\section{ReWAM}

\subsection{Overview}

ReWAM jointly learns an action policy and a predictor of future visual representations. Given the previous and current sampled multi-view observations $o_{-1}$ and $o_0$, proprioception $s_0$, and a language instruction $\ell$, the policy predicts an action chunk $a_{1:H}$. Here, $H$ denotes the shared future time horizon, covered by actions and visual states at their respective sampling rates. With observations indexed at the image-sampling rate, ReWAM encodes non-overlapping pairs as
\begin{equation}
    z_j=E_\phi(o_{j-1},o_{j}).
\end{equation}
The encoder $E_\phi$ combines a frozen DINOv3-L backbone, feature calibration, and a trainable TRB with parameters $\phi$. The current state $z_0$ conditions both branches, and $z_{1:H}$ provide the world-prediction targets. Sampling and execution details appear in Appendix~\ref{app:implementation}.

Following Fast-WAM~\citep{Fast-WAM}, ReWAM uses a Mixture-of-Transformers (MoT) architecture with world and action Diffusion Transformer (DiT) branches, denoted by $W_\theta$ and $A_\theta$. Both branches receive conditioning embeddings $c$ of $\ell$ and $s_0$. Future-state and action tokens attend to current-state world features, but attention between the two token groups is blocked. The world branch uses causal temporal attention and independently sampled noise levels for future states, while the action branch predicts the entire action chunk at once. The conditional model therefore factorizes as
\begin{equation}
    p_\theta(z_{1:H},a_{1:H}\mid z_0,c)
    =p_\theta(z_{1:H}\mid z_0,c)\,p_\theta(a_{1:H}\mid z_0,c),
\end{equation}
where $\theta$ denotes the trainable WAM parameters. Both branches are trained with flow matching. At inference, the world branch processes $z_0$ once to condition action generation, without generating future states.

\subsection{Feature Calibration}

Following RAE~\citep{RAE,RAEv2}, we calibrate DINO representations through multi-layer aggregation, normalization, and a representation-aware noise schedule.

\paragraph{Multi-layer aggregation.}
For each image and camera view, we extract patch features from DINOv3-L blocks $\mathcal{S}=\{12,14,16,18,20,22,24\}$, using one-based indexing. Let $F^{(l)}\in\mathbb{R}^{N\times d}$ denote the features from block $l$ after non-affine LayerNorm and removal of special tokens, where $N$ is the number of patches in that view and $d=1024$. We average the selected layers and add the spatial mean of the final layer to every patch:
\begin{equation}
    F_p^{\mathrm{MLA}}
    =\frac{1}{|\mathcal{S}|}\sum_{l\in\mathcal{S}}F^{(l)}_p
    +\frac{1}{N}\sum_{q=1}^{N}F^{(24)}_q,
    \qquad p=1,\ldots,N.
\end{equation}
This combines features across encoder depths with image-level context while preserving the patch grid and feature width.

\paragraph{Normalization.}
For the calibrated frozen-feature baseline in \cref{tab:feat_calib}, we standardize prediction targets using mean and variance estimates fixed before WAM training. In ReWAM, the TRB changes the prediction space during training, so we normalize its outputs with running statistics, as detailed in \cref{sec:trb}.

\paragraph{Representation-aware noise schedule.}
Let $\tau\in[0,1]$ denote flow time, with $\tau=0$ corresponding to clean data and $\tau=1$ to Gaussian noise. Let $D$ be the total number of scalar elements in the future prediction target, excluding the current state. Following RAE, we sample $u\sim\mathcal{U}(0,1)$ and set
\begin{equation}
    \tau=\frac{\alpha u}{1+(\alpha-1)u},
    \qquad \alpha=\sqrt{D/D_{\mathrm{base}}}.
\end{equation}
We set $D_{\mathrm{base}}=4096$. The shift accounts for both token count and channel width, favoring noisier inputs for larger targets. We sample a separate flow time for each future state using the same shift, while the current state remains clean.

\subsection{Temporal Representation Bottleneck}
\label{sec:trb}

The TRB $B_\phi$ converts frame-wise DINO features into compact world states. For each camera, we flatten non-overlapping $2\times2\times2$ spatiotemporal blocks into $8d$-dimensional tokens. We concatenate tokens across views, then apply multi-head self-attention and a linear projection to obtain $128$-dimensional tokens. Before normalization, the state is
\begin{equation}
    z_j^{\mathrm{raw}}=B_\phi(\operatorname{Patchify}\!\left(F_{j-1}^{\mathrm{MLA}},F_{j}^{\mathrm{MLA}}\right)).
\end{equation}

\paragraph{Causal running normalization.}
We normalize TRB outputs using per-channel running means $\mu$ and variances $\sigma^2$ accumulated from preceding batches:
\begin{equation}
    z_j=\frac{z_j^{\mathrm{raw}}-\mu}{\sqrt{\sigma^2+\varepsilon}},
\end{equation}
where $\varepsilon$ is a numerical stabilizer. We then update $\mu$ and $\sigma^2$ with the current batch moments using an exponential moving average. The statistics remain fixed at inference.

\subsection{Action-Grounded Representation Shaping}

AGRS assigns distinct roles to the two objectives: the action objective shapes the TRB, and the future prediction objective trains the world model in the resulting representation space. Only action-loss gradients through the current state $z_0$ reach the TRB.

\paragraph{Joint world-action learning.}
We use the linear flow-matching formulation of Fast-WAM. For a clean target $y$ and Gaussian noise $\epsilon\sim\mathcal{N}(0,I)$, we construct
\begin{equation}
    y^{\tau}=(1-\tau)y+\tau\epsilon.
\end{equation}
Here, $y$ is a future state or an action chunk, and the target velocity is $\epsilon-y$. At inference, we integrate the predicted action velocity from noise at $\tau=1$ to data at $\tau=0$.

Let $z=z_{1:H}$, $a=a_{1:H}$, and $\boldsymbol{\tau}_z=(\tau_{z,1},\ldots,\tau_{z,H})$. A bar denotes stop-gradient, and noising is applied independently to each future state. The world and action velocity predictors are trained with
\begin{align}
    \mathcal{L}_{\mathrm{world}}
     & =\mathbb{E}\!\left[
                        \left\|W_\theta(\bar z^{\boldsymbol{\tau}_z},\boldsymbol{\tau}_z;\bar z_0,c)
                        -(\epsilon_z-\bar z)\right\|_2^2
                        \right],
    \\
    \mathcal{L}_{\mathrm{action}}
     & =\mathbb{E}\!\left[
                        \left\|A_\theta(a^{\tau_a},\tau_a;z_0,c)
                        -(\epsilon_a-a)\right\|_2^2
                        \right].
\end{align}

\paragraph{Asymmetric gradient routing.}
We jointly train the WAM with both objectives, while routing only action-loss gradients to the TRB:
\begin{equation}
    \nabla_\theta\mathcal{L}
    =\nabla_\theta\mathcal{L}_{\mathrm{world}}
    +\nabla_\theta\mathcal{L}_{\mathrm{action}},
    \qquad
    \nabla_\phi\mathcal{L}
    =\left(\frac{\partial z_0}{\partial\phi}\right)^{\!\top}
    \frac{\partial\mathcal{L}_{\mathrm{action}}}{\partial z_0}.
\end{equation}
World-loss gradients stop at both the current and future representations. Action-loss gradients pass through the world branch's current-state features and $z_0$ into the TRB. Thus, the world branch learns from both objectives, but the representation encoder is shaped only by the action objective.

\begin{table}[!t]
    \centering
    \begin{minipage}{\linewidth}
    \centering
    \caption{\textbf{Evaluation results on RoboTwin 2.0.} Success rates (\%) are averaged across 50 tasks, with 100 evaluation episodes per task in each setting. Embodied PT indicates embodied pre-training.}
    \label{tab:robotwin}
    \begin{tabular}{@{}lcccc@{}}
        \toprule
        Method                                     & Embodied PT                                                                   & Clean & Rand & Avg  \\
        \midrule
        $\pi_0$~\scriptsize{\citep{Pi_0}}          & \cmark                                                                        & 65.9  & 58.4 & 62.2 \\
        $\pi_{0.5}$~\scriptsize{\citep{Pi_0.5}}    & \cmark                                                                        & 82.7  & 76.8 & 79.8 \\
        Motus~\scriptsize{\citep{Motus}}           & \cmark                                                                        & 88.7  & 87.0 & 87.8 \\
        Fast-WAM~\scriptsize{\citep{Fast-WAM}}     & \xmark                                                                        & 91.9  & 91.8 & 91.8 \\
        LingBot-VA~\scriptsize{\citep{LingBot-VA}} & \cmark                                                                        & 92.9  & 91.5 & 92.2 \\
        AHA-WAM~\scriptsize{\citep{AHA-WAM}}       & \xmark                                                                        & 93.4  & 92.2 & 92.8 \\
        \multicolumn{1}{@{}>{\columncolor{gray!10}[0pt][\tabcolsep]}l}{\textbf{ReWAM \scriptsize{(Ours)}}}
                                                   & \multicolumn{1}{>{\columncolor{gray!10}}c}{\xmark}
                                                   & \multicolumn{1}{>{\columncolor{gray!10}}c}{\textbf{93.6}}
                                                   & \multicolumn{1}{>{\columncolor{gray!10}}c}{\textbf{93.6}}
                                                   & \multicolumn{1}{>{\columncolor{gray!10}[\tabcolsep][0pt]}c@{}}{\textbf{93.6}}                       \\
        \bottomrule
    \end{tabular}
\end{minipage}

    \par\vspace{1.4em}
    \begin{minipage}{\linewidth}
    \centering
    \caption{\textbf{Evaluation results on RoboDojo.} SR denotes success rate (\%). Fast-WAM$^{*}$ uses the same embodied pre-training as ReWAM. Within each pre-training setting, the best results are in \textbf{bold} and the second-best are \underline{underlined}.}
    \label{tab:robodojo}
    \newcommand{\metricwidth}{2.7em}
    \newcommand{\withingroupgap}{0.3em}
    \newcommand{\betweengroupgap}{0.5em}
    \setlength{\tabcolsep}{0pt}
    \newcolumntype{R}{>{\raggedleft\arraybackslash}p{\metricwidth}}
    \resizebox{\linewidth}{!}{%
        \begin{tabular}{@{}l@{\hspace{\betweengroupgap}}
            *{5}{R@{\hspace{\withingroupgap}}R@{\hspace{\betweengroupgap}}}
            R@{\hspace{\withingroupgap}}R@{}}
            \toprule
            \multirow{2}{*}{Method}                            & \multicolumn{2}{c}{Generalization} & \multicolumn{2}{c}{Precision} & \multicolumn{2}{c}{Long-Horizon} & \multicolumn{2}{c}{Memory} & \multicolumn{2}{c}{Open} & \multicolumn{2}{c}{Average}                                                                                                                    \\
            \cmidrule(lr){2-3}\cmidrule(lr){4-5}\cmidrule(lr){6-7}\cmidrule(lr){8-9}\cmidrule(lr){10-11}\cmidrule(lr){12-13}
                                                               & Score                              & SR                            & Score                            & SR                         & Score                    & SR                          & Score            & SR               & Score            & SR               & Score             & SR               \\
            \midrule
            \rowcolor{gray!10}
            \multicolumn{13}{l}{\textbf{\textit{w/ Embodied Pre-training}}}                                                                                                                                                                                                                                                                                                     \\
            LDA-1B~\tiny{\citep{LDA-1B}}                       & 0.71                               & 0.17                          & 3.21                             & 0.50                       & 1.92                     & 0.08                        & 2.08             & 1.78             & 0.00             & 0.00             & 1.58              & 0.51             \\
            InternVLA-A1~\tiny{\citep{InternVLA-A1}}           & 2.87                               & 1.83                          & 3.00                             & 0.92                       & 4.79                     & 1.17                        & 1.58             & 1.33             & 0.17             & 0.17             & 2.48              & 1.08             \\
            GR00T-N1.7~\tiny{\citep{GR00T}}                    & 2.16                               & 1.22                          & 2.54                             & 0.67                       & 8.30                     & 3.58                        & 1.06             & 0.89             & 0.18             & 0.17             & 2.85              & 1.31             \\
            $\pi_0$~\tiny{\citep{Pi_0}}                        & 3.94                               & 2.56                          & 3.56                             & 0.75                       & 6.19                     & 2.00                        & 3.47             & 2.11             & 0.25             & 0.25             & 3.48              & 1.53             \\
            LingBot-VLA~\tiny{\citep{LingBot-VLA}}             & 6.71                               & 4.28                          & 5.33                             & 1.83                       & 10.89                    & 5.25                        & 3.82             & 2.78             & 0.72             & 0.67             & 5.50              & 2.96             \\
            Galaxea-G0-VLA~\tiny{\citep{Galaxea_G0}}           & 4.53                               & 2.83                          & 8.10                             & 3.83                       & 12.60                    & 5.58                        & 3.17             & 1.89             & 0.70             & 0.67             & 5.82              & 2.96             \\
            Xiaomi-Robotics-0~\tiny{\citep{Xiaomi-Robotics-0}} & 7.43                               & 5.56                          & 8.42                             & 4.58                       & 13.51                    & 6.92                        & 5.07             & 3.67             & 0.22             & 0.17             & 6.93              & 4.18             \\
            X-VLA~\tiny{\citep{X-VLA}}                         & 10.48                              & 6.78                          & \textbf{18.32}                   & \textbf{12.00}             & 16.53                    & 9.75                        & 4.76             & 3.56             & 0.55             & 0.50             & 10.13             & 6.52             \\
            $\pi_{0.5}$~\tiny{\citep{Pi_0.5}}                  & 13.37                              & 8.17                          & 12.40                            & 5.50                       & 23.54                    & 14.67                       & 5.78             & 4.56             & \textbf{1.98}    & \textbf{1.67}    & 11.41             & 6.91             \\
            Spatial Forcing~\tiny{\citep{Spatial_Forcing}}     & \underline{14.12}                  & \underline{9.33}              & \underline{17.33}                & \underline{10.58}          & 23.26                    & 14.58                       & 5.43             & 4.11             & \underline{1.78} & \underline{1.58} & \underline{12.38} & 8.04             \\
            HyVLA-0.5~\tiny{\citep{Hy-Embodied-0.5-VLA}}       & 11.77                              & 8.39                          & 13.81                            & 8.00                       & \textbf{25.74}           & \underline{14.92}           & \textbf{13.37}   & \textbf{12.11}   & 0.65             & 0.58             & \textbf{13.07}    & \textbf{8.80}    \\
            Fast-WAM$^{*}$~\tiny{\citep{Fast-WAM}}             & 12.80                              & 9.17                          & 12.32                            & 4.25                       & 22.00                    & 14.50                       & 6.65             & 5.33             & 0.95             & 0.75             & 10.95             & 6.80             \\
            \textbf{ReWAM \scriptsize{(Ours)}}                 & \textbf{14.65}                     & \textbf{10.50}                & 12.32                            & 6.75                       & \underline{25.20}        & \textbf{16.25}              & \underline{9.05} & \underline{7.67} & 0.25             & 0.25             & 12.29             & \underline{8.28} \\
            \midrule
            \rowcolor{gray!10}
            \multicolumn{13}{l}{\textbf{\textit{w/o Embodied Pre-training}}}                                                                                                                                                                                                                                                                                                    \\
            Fast-WAM~\tiny{\citep{Fast-WAM}}                   & 2.34                               & 1.11                          & 1.96                             & 0.00                       & 9.14                     & 5.17                        & 3.55             & 3.44             & 0.42             & 0.42             & 3.48              & 2.03             \\
            AHA-WAM~\tiny{\citep{AHA-WAM}}                     & 5.79                               & 3.28                          & 5.86                             & 2.42                       & 8.61                     & 2.67                        & 2.97             & 2.78             & \textbf{0.88}    & \textbf{0.83}    & 4.82              & 2.39             \\
            GigaWorld-Policy~\tiny{\citep{GigaWorld-Policy}}   & 5.34                               & 2.89                          & 6.15                             & 1.83                       & 15.51                    & 8.92                        & 3.46             & 2.22             & 0.54             & 0.50             & 6.20              & 3.27             \\
            StarVLA-$\alpha$~\tiny{\citep{StarVLA}}            & 3.93                               & 2.33                          & \underline{9.90}                 & \underline{4.33}           & 14.15                    & 6.50                        & 3.34             & 2.44             & \underline{0.68} & \underline{0.58} & 6.40              & 3.24             \\
            X-WAM~\tiny{\citep{X-WAM}}                         & \underline{7.39}                   & \underline{3.33}              & 6.72                             & 1.83                       & \underline{17.47}        & \underline{9.08}            & \underline{6.32} & \underline{4.67} & 0.57             & 0.25             & \underline{7.69}  & \underline{3.83} \\
            \textbf{ReWAM \scriptsize{(Ours)}}                 & \textbf{7.52}                      & \textbf{4.33}                 & \textbf{13.30}                   & \textbf{7.50}              & \textbf{18.90}           & \textbf{11.00}              & \textbf{7.00}    & \textbf{6.00}    & 0.38             & 0.25             & \textbf{9.42}     & \textbf{5.82}    \\
            \bottomrule
        \end{tabular}%
    }
\end{minipage}

\end{table}

\section{Experiments}

\subsection{Setup}

\paragraph{Implementation Details.}
ReWAM uses frozen DINOv3 ViT-L/16~\citep{DINOv3} features and UMT5 embeddings of language instructions. The world and action DiT branches each contain 30 layers, with hidden widths of $2048$ and $1024$, respectively, and approximately 3.58B parameters in total. Head-camera images are resized to $256\times320$ pixels and wrist-camera images to $128\times160$ pixels. Actions and proprioception are represented as 14-dimensional joint-and-gripper vectors and standardized using z-score normalization. We train with AdamW on 32 H20 GPUs, with a batch size of 512 and a peak learning rate of $10^{-4}$ that decays to $10^{-6}$ on a cosine schedule.

\paragraph{RoboTwin 2.0.}
We train a multi-task policy on the 50-task RoboTwin 2.0 benchmark~\citep{RoboTwin} using 2,500 clean trajectories and 25,000 trajectories from the random setting. The main comparison uses five training epochs, aligned with the baseline protocol. Representation comparisons and ablations, including the Fast-WAM baselines, use a matched one-epoch training budget. We evaluate 100 episodes per task in each of the clean and random settings and average success rates across tasks.

\paragraph{RoboDojo.}
We evaluate on the simulation suite of RoboDojo~\citep{RoboDojo}, which covers five capability categories: Generalization, Precision, Long-Horizon, Memory, and Open. The training set contains 3,500 trajectories across 35 tasks. We train ReWAM for 50k steps with and without prior embodied pre-training. We report the benchmark's partial-progress score and binary success rate, averaging each metric across the five categories and comparing methods within each pre-training setting.

\paragraph{Embodied Pre-training.}
To assess the benefit of embodied pre-training with additional data, we pre-train ReWAM and Fast-WAM for 75k steps on the same 600 hours of self-collected real-robot manipulation demonstrations. The dataset covers a broad range of manipulation tasks, with collection environments and camera settings that differ from those used in our downstream real-robot experiments.

\paragraph{Real-Robot Evaluation.}
We evaluate ReWAM on a bimanual Piper platform across three tasks: Collect Objects, Fold Clothes, and Unpack Lunchbox. After embodied pre-training, ReWAM and Fast-WAM each undergo 10k steps of task-specific post-training on demonstrations collected in the target scene. We report success rates over 20 trials per task.

\subsection{Main Results}

ReWAM achieves 93.6\% success in both settings of RoboTwin 2.0 (\cref{tab:robotwin}), exceeding the strongest compared method by $0.2$ percentage points on clean and $1.4$ points on random. Its performance in the random setting indicates robustness to the benchmark's scene variations without embodied pre-training.

On RoboDojo, ReWAM leads the compared methods without embodied pre-training, with an average score of $9.42$ and a success rate of $5.82\%$ (\cref{tab:robodojo}). Embodied pre-training raises these results to $12.29$ and $8.28\%$, highlighting ReWAM's potential to improve further with larger-scale embodied training data. In this setting, ReWAM achieves the highest Generalization score and success rate and the highest Long-Horizon success rate.

The benefits of embodied pre-training vary across capabilities. Generalization and Long-Horizon success rates rise from $4.33\%$ to $10.50\%$ and from $11.00\%$ to $16.25\%$, respectively. ReWAM remains competitive on Precision tasks without explicit geometric enhancements, though it trails geometry-enhanced methods such as X-VLA~\cite{X-VLA} and Spatial Forcing~\cite{Spatial_Forcing}. ReWAM also falls substantially behind the leading methods in Open instruction following. We attribute the gap in open instruction following to the absence of generative video pre-training, which limits instruction generalization.

\subsection{Representation Design and Ablations}

\begin{table}[t]
    \centering
    \captionsetup{justification=raggedright,singlelinecheck=false,skip=6pt}
    \newcommand{\ablationleftshare}{0.56}
    \newlength{\ablationcolumngap}
    \setlength{\ablationcolumngap}{0.075\linewidth}
    \newlength{\ablationavailablewidth}
    \newlength{\ablationleftwidth}
    \newlength{\ablationrightwidth}
    \newlength{\ablationpanelwidth}
    \newlength{\ablationcolumnheight}
    \setlength{\ablationavailablewidth}{\dimexpr\linewidth-\ablationcolumngap\relax}
    \setlength{\ablationleftwidth}{\ablationleftshare\ablationavailablewidth}
    \setlength{\ablationrightwidth}{\dimexpr\ablationavailablewidth-\ablationleftwidth\relax}
    \newsavebox{\featcalibtablebox}
    \newsavebox{\tempbtnktablebox}
    \newsavebox{\agrsablationtablebox}
    \newsavebox{\agrsdimstablebox}

    \newcommand{\ablationtable}[3]{%
        \begin{minipage}[t]{\ablationpanelwidth}
            \vspace{0pt}%
            \caption{#1}\label{#2}
            \setlength{\tabcolsep}{3pt}
            #3\par
        \end{minipage}%
    }
    \newcommand{\storeablationtable}[3]{%
        \setlength{\ablationpanelwidth}{#3}%
        \begin{lrbox}{#1}%
            \input{#2}\unskip
        \end{lrbox}%
    }
        \setlength{\ablationpanelwidth}{\ablationleftwidth}%
        \begin{lrbox}{\featcalibtablebox}%
            \ablationtable{\textbf{Feature calibration.} Gen PT denotes generative video pre-training; ``+'' rows are cumulative.}{tab:feat_calib}{%
    \begin{tabular*}{\linewidth}{@{\extracolsep{\fill}}lcrr@{}}
        \toprule
        Model                                            & Gen PT         & Clean          & Rand  \\
        \midrule
        Fast-WAM                                         & \cmark         & 87.78          & 86.43 \\
        Fast-WAM                                         & \xmark         & 66.10          & 62.32 \\
        \midrule
        Raw DINO WAM                                     & \xmark         & 71.00          & 71.54 \\
        \multicolumn{2}{l}{+ Normalization}              & 76.10          & 74.28                  \\
        \multicolumn{2}{l}{+ Repr.-aware noise schedule} & 77.38          & 75.82                  \\
        \multicolumn{2}{l}{+ Multi-layer aggregation}    & \textbf{83.90} & \textbf{83.03}         \\
        \bottomrule
    \end{tabular*}%
}
\unskip
        \end{lrbox}%

        \setlength{\ablationpanelwidth}{\ablationrightwidth}%
        \begin{lrbox}{\tempbtnktablebox}%
            \ablationtable{\textbf{Temporal Representation Bottleneck design.}}{tab:temp_btnk}{%
    \begin{tabular*}{\linewidth}{@{\extracolsep{\fill}}lrr@{}}
        \toprule
        TRB variant             & Clean          & Rand           \\
        \midrule
        No TRB                  & 83.90          & 83.03          \\
        Fixed random projection & 86.76          & 85.94          \\
        RGB recon. codec        & 71.04          & 70.48          \\
        DINO recon. codec       & 85.58          & 86.08          \\
        AGRS                    & \textbf{90.70} & \textbf{89.12} \\
        \bottomrule
    \end{tabular*}%
}
\unskip
        \end{lrbox}%

        \setlength{\ablationpanelwidth}{\ablationleftwidth}%
        \begin{lrbox}{\agrsablationtablebox}%
            \ablationtable{\textbf{AGRS gradient routing.} Checks mark gradients reaching the TRB.}{tab:agrs_routing}{%
    \begin{tabular*}{\linewidth}{@{\extracolsep{\fill}}cccrr@{}}
        \toprule
        \multicolumn{2}{@{}c}{World$\to$TRB} & Action$\to$TRB & \multirow{2}{*}{Clean} & \multirow{2}{*}{Rand}                  \\
        \cmidrule(lr){1-2}\cmidrule(lr){3-3}
        Current                              & Future         & Current                &                       &                \\
        \midrule
        \cmark                               & \cmark         & \cmark                 & 79.40                 & 77.40          \\
        \cmark                               & \xmark         & \cmark                 & 84.32                 & 82.90          \\
        \xmark                               & \xmark         & \cmark                 & \textbf{90.70}        & \textbf{89.12} \\
        \bottomrule
    \end{tabular*}%
}
\unskip
        \end{lrbox}%

        \setlength{\ablationpanelwidth}{\ablationrightwidth}%
        \begin{lrbox}{\agrsdimstablebox}%
            \ablationtable{\textbf{TRB dimensionality.}}{tab:agrs_dims}{%
    \begin{tabular*}{\linewidth}{@{\extracolsep{\fill}}lcrr@{}}
        \toprule
        TRB variant                                                 & Dim & Clean          & Rand           \\
        \midrule
        \multirow{3}{*}{\shortstack[l]{Fixed random \\ projection}} & 512 & 86.76          & 85.94          \\
                                                                    & 256 & 86.09          & 85.18          \\
                                                                    & 128 & 84.26          & 81.87          \\
        \midrule
        \multirow{3}{*}{AGRS}                                       & 512 & 89.08          & 87.64          \\
                                                                    & 128 & \textbf{90.70} & \textbf{89.12} \\
                                                                    & 64  & 89.90          & 89.08          \\
        \bottomrule
    \end{tabular*}%
}
\unskip
        \end{lrbox}%

    \setlength{\ablationcolumnheight}{\dimexpr
        \ht\featcalibtablebox+\dp\featcalibtablebox+
        \ht\agrsablationtablebox+\dp\agrsablationtablebox\relax}
    \ifdim\dimexpr\ht\tempbtnktablebox+\dp\tempbtnktablebox+
        \ht\agrsdimstablebox+\dp\agrsdimstablebox\relax>\ablationcolumnheight
        \setlength{\ablationcolumnheight}{\dimexpr
            \ht\tempbtnktablebox+\dp\tempbtnktablebox+
            \ht\agrsdimstablebox+\dp\agrsdimstablebox\relax}
    \fi
    \addtolength{\ablationcolumnheight}{1.4em}
    \begin{minipage}[t][\ablationcolumnheight][s]{\ablationleftwidth}
        \usebox{\featcalibtablebox}
        \par\nointerlineskip\vfill
        \usebox{\agrsablationtablebox}
        \par\vspace{0pt}
    \end{minipage}\hspace{\ablationcolumngap}%
    \begin{minipage}[t][\ablationcolumnheight][s]{\ablationrightwidth}
        \usebox{\tempbtnktablebox}
        \par\nointerlineskip\vfill
        \usebox{\agrsdimstablebox}
        \par\vspace{0pt}
    \end{minipage}
\end{table}

\paragraph{Reconstruction Fidelity versus Control Utility.}
We compare Video-VAE and DINO WAMs in \cref{tab:feat_calib}. Removing generative video pre-training reduces Fast-WAM's success by $21.68/24.11$ percentage points on clean/random. Without this pre-training, raw DINO features outperform Video-VAE latents by $4.90/9.22$ points, despite lower RGB reconstruction SSIM ($0.70$ versus $0.91$). An RGB reconstruction codec likewise yields lower success than a DINO reconstruction codec (\cref{tab:temp_btnk}).  Reconstruction fidelity thus does not necessarily ensure a representation's utility for control.

\paragraph{Feature Calibration.}
Progressively calibrating DINO features with normalization, a representation-aware noise schedule, and multi-layer aggregation yields successive gains of $3.92$, $1.41$, and $6.87$ percentage points in average success, as detailed in \cref{tab:feat_calib}. Together, these components improve average success by $12.20$ points over raw DINO features. The gains indicate that pre-trained perceptual features benefit from calibration to the distribution and structure required by diffusion-based dynamics modeling.

\paragraph{Temporal Representation Bottleneck.}
To assess temporal organization, we compare the representation designs in \cref{tab:temp_btnk}. Fixed random projection concatenates features at each spatial location across a temporal window and projects them to $512$ dimensions. This simple construction improves average success by $2.89$ percentage points over the calibrated frame-wise baseline, highlighting the value of temporal organization. The result is consistent with prior work on fixed random projections of pre-trained DINO features as prediction latents~\citep{MIRA}. We also pre-train the TRB as a codec to reconstruct RGB images or DINO features and freeze it during WAM training. The RGB reconstruction codec substantially reduces success, whereas the DINO reconstruction codec falls slightly below fixed random projection on average, suggesting that reconstruction-based training alone is insufficient to improve these temporal representations for control.

\paragraph{Action-Grounded Representation Shaping.}
AGRS achieves $90.70\%/89.12\%$ success on clean/random, outperforming fixed random projection by $3.94/3.18$ percentage points in \cref{tab:temp_btnk}. The gradient-routing ablation in \cref{tab:agrs_routing} supports keeping world-loss gradients out of the TRB: allowing them through the current-representation path reduces average success by $6.30$ percentage points, and allowing them through both current and future paths degrades performance further. These results suggest that letting the world objective reshape its prediction targets may encourage representation collapse or predictive shortcuts.
The dimensionality comparison in \cref{tab:agrs_dims} highlights a further advantage of AGRS. Reducing fixed random projection from $512$ to $128$ dimensions lowers average success by $3.29$ percentage points. AGRS performs best among the tested widths at $128$ dimensions and remains effective at $64$ dimensions. This contrast suggests that action grounding learns more effective control representations, enabling strong performance even at substantially lower dimensionality.

\begin{figure}[tp]
    \centering
    \input{figures/2_action_aligned_tsne.tex}
    \par\vspace{1.4em}
    \input{figures/3_real_robot.tex}
\end{figure}

\subsection{Action-Related Structure in Representations}

We investigate the action-related structure in ReWAM and Video-VAE representations using t-SNE and nearest-neighbor direction agreement. \Cref{fig:action_aligned_tsne}(a) shows independent t-SNE projections of commanded end-effector trajectories and representation variations for the same transitions: one from each of 242 left-arm episodes per task, selected from demonstrations in the random setting. Colors indicate the dominant direction of commanded displacement along the robot's coordinate axes. Highlighted subsets are balanced across directions by displacement magnitude, while the remaining transitions appear in gray. AGRS exhibits clearer local grouping by direction, suggesting that its representation variations capture action-related information.

We also measure whether nearby representation variations correspond to the same motion direction in the original representation space. We find each highlighted transition's ten nearest neighbors within the balanced subset, excluding itself, and compute the fraction sharing its commanded-motion direction. \Cref{fig:action_aligned_tsne}(b) reports the mean agreement, which is higher in AGRS space than in Video-VAE space.

\subsection{Real-Robot Evaluation}

We assess ReWAM on physical hardware across three tasks. Collect Objects requires picking up tabletop items and placing them in a basket; Fold Clothes involves manipulating a garment; and Unpack Lunchbox requires opening a lunchbox and retrieving food. Both methods complete all 20 Collect Objects trials, while ReWAM improves success by 15 percentage points on Fold Clothes and 20 points on Unpack Lunchbox (\cref{fig:real_robot}). Across the three tasks, ReWAM succeeds in 55 of 60 trials (91.7\%), compared with 48 of 60 (80.0\%) for Fast-WAM. These gains highlight ReWAM's performance advantage on more challenging tasks involving deformable objects and multi-stage manipulation. We also observe autonomous retries, despite their absence from the task demonstrations. Additional real-robot demonstrations are available in the supplementary materials.

\section{Conclusion}

Our study shows that effective world-action modeling depends on how its representation space is constructed and adapted. Reconstruction fidelity alone does not determine control performance, and pre-trained perceptual features benefit from calibration and temporal organization. ReWAM combines these components with action-grounded representation shaping, letting the policy determine what the representation encodes, while the world model learns its dynamics. Results on RoboTwin 2.0, RoboDojo, and real-robot manipulation demonstrate the effectiveness of this design without generative video pre-training. These findings highlight the potential of representation design to advance joint prediction and control.






\bibliography{references}
\bibliographystyle{unsrtnat}

\clearpage
\appendix
\section{Implementation and Evaluation Details}
\label{app:implementation}

\paragraph{Temporal sampling and horizons.}
Each training clip contains ten multi-view observations sampled every $8$ control steps, at offsets $[-8,0,8,16,24,32,40,48,56,64]$. The first pair defines $z_0$, and the remaining four pairs define $z_{1:4}$ over the $64$-action horizon. Inference requires only the historical and current observations. At the first query, the current observation substitutes for the missing history. We apply the same substitution during RoboTwin training with probability $0.1$. Padded targets are excluded from the corresponding losses.

\paragraph{TRB architecture.}
The head camera yields a $16\times20$ DINO patch grid, and each wrist camera yields an $8\times10$ grid. Spatiotemporal patchification produces $80+20+20=120$ tokens per state, each initially $8192$-dimensional. A single eight-head attention layer mixes tokens across views within each temporal pair. It uses two-dimensional RoPE with base $10{,}000$, coordinates scaled by $16$, and learned camera-specific query and key biases.

\paragraph{Inference and action execution.}
At each query, we cache the world branch's current-state keys and values, generate actions from Gaussian noise using Euler updates with decreasing flow time, and execute a prefix of the predicted chunk. \Cref{tab:inference_settings} lists the execution and denoising budgets used for evaluation.

\begin{table}[htbp]
    \centering
    \caption{\textbf{ReWAM inference settings.} Executed actions are the prefix applied before the next policy query.}
    \label{tab:inference_settings}
    \begin{tabular}{@{}lccc@{}}
        \toprule
        Setting          & Predicted actions & Executed actions & Denoising steps \\
        \midrule
        RoboTwin 2.0     & 64                & 24               & 10              \\
        RoboDojo         & 64                & 10               & 20              \\
        Piper real robot & 64                & 32               & 20              \\
        \bottomrule
    \end{tabular}
\end{table}

\section{Embodied Pre-training Data Scale}

\Cref{tab:embodied_pretraining_data} compares the reported durations of embodied pre-training datasets for selected methods evaluated on RoboDojo. ReWAM uses approximately $600$ hours of self-collected robot demonstrations, substantially less than the amounts reported for the listed baselines. With this pre-training, ReWAM achieves an average score of $12.29$ and a success rate of $8.28\%$, as shown in \cref{tab:robodojo}. Our matched Fast-WAM baseline uses the same $600$-hour dataset. Differences in data composition should be considered when interpreting the comparison.

\begin{table}[htbp]
    \centering
    \caption{\textbf{Embodied pre-training data scale for selected RoboDojo methods.} Hours measure the duration of embodied training data. A ``+'' indicates a reported lower bound, and $\approx$ denotes an approximate duration.}
    \label{tab:embodied_pretraining_data}
    \begin{tabular}{@{}lr@{}}
        \toprule
        Method                                & Data duration in hours \\
        \midrule
        Galaxea-G0-VLA~\citep{Galaxea_G0}     & $2{,}000+$             \\
        $\pi_0$~\citep{Pi_0}                  & $10{,}000+$            \\
        $\pi_{0.5}$~\citep{Pi_0.5}            & $10{,}000+$            \\
        HyVLA-0.5~\citep{Hy-Embodied-0.5-VLA} & $10{,}000+$            \\
        LingBot-VLA~\citep{LingBot-VLA}       & $\approx 20{,}000$     \\
        \midrule
        \textbf{ReWAM}                        & $\approx 600$          \\
        \bottomrule
    \end{tabular}
\end{table}

\section{Contribution of World Co-training}
\label{app:world_loss_ablation}
To isolate the contribution of world co-training, we disable the world loss in the raw DINO WAM configuration from \cref{tab:feat_calib} while keeping the backbone and architecture unchanged. As shown in \cref{tab:world_loss_ablation}, world co-training improves clean/random success by $2.60/2.99$ percentage points, providing gains beyond those from pre-trained DINO features alone.

\begin{table}[htbp]
    \centering
    \caption{\textbf{World-loss ablation for DINO WAM on RoboTwin 2.0.}}
    \label{tab:world_loss_ablation}
    \begin{tabular}{@{}lrr@{}}
        \toprule
        Method       & Clean & Rand  \\
        \midrule
        DINO policy  & 68.40 & 68.55 \\
        Raw DINO WAM & 71.00 & 71.54 \\
        \bottomrule
    \end{tabular}
\end{table}

\section{Detached Pixel-Reconstruction Probe}
\label{app:reconstruction_probe}

The SSIM probe measures how much pixel information can be recovered from the representation. During training, an auxiliary decoder $D_\psi(\bar z_j)$ reconstructs future RGB observation pairs from detached latents. The reconstruction loss updates only the decoder parameters $\psi$, without affecting the encoder or WAM. The decoder is not used at inference.

\section{Representation Analysis Protocol}
\label{app:representation_analysis}

\paragraph{Data source and transition selection.}
We examine all $500$ RoboTwin demonstrations from the random setting for each displayed task. We retain episodes in which the left gripper's command range exceeds $0.1$ and the right gripper's does not. Candidate transitions start at $16,24,32,\ldots$ and span $32$ control steps. Throughout each window, including both endpoints, the active gripper's command and recorded state must remain below $0.2$ and $0.25$, respectively, in dataset units. We select the middle eligible start in each episode, using the later of the two middle starts when the count is even. This yields $242$ transitions each for Move Stapler Pad and Place Mouse Pad. Selection uses only action and state records, independently of representation features or projections.

\paragraph{Direction balancing.}
We assign each transition to the signed axis with the largest absolute endpoint displacement. To qualify for highlighting, a transition must have a displacement of at least $0.04$ m within $45^\circ$ of that axis. Directions with fewer than $20$ eligible candidates are excluded. Within displacement-magnitude bins $[0.10,0.14)$, $[0.14,0.18)$, $[0.18,0.22)$, and $[0.22,0.30)$ m, we sample equal counts per direction without replacement. We omit any bin with fewer than three candidates in a retained direction. This yields $144$ highlighted transitions for Move Stapler Pad and $168$ for Place Mouse Pad, with $36$ and $42$ per direction, respectively. Both subsets cover $+X$, $-X$, $+Y$, and $-Y$.

\paragraph{t-SNE projection.}
Each t-SNE fit includes all $242$ transitions, with those outside the highlighted subset shown in gray. We use Euclidean distances, perplexity $30$, PCA initialization, an automatic learning rate, and $1500$ iterations. We do not reduce feature dimensionality with PCA or standardize features before fitting. Projections are fitted independently, so orientations and absolute positions cannot be compared across panels.

\end{document}